\documentclass{article}
\usepackage{imav}
\usepackage{amsmath}
\usepackage{times}
\usepackage{graphicx}
\usepackage{hyperref}
\usepackage{wasysym}

\title{Real-time and Autonomous Detection of Helipad for Landing Quad-Rotors by Visual Servoing}
\author{Archit Rungta\thanks{Email address: archit120@iitkgp.ac.in}, Yash Soni\thanks{Email address: yashsoni501@iitkgp.ac.in}, Parakh Agarwal, Biswajit Ghosh, Somesh Kumar \\ Indian Institute of Technology, Kharagpur, India}

\begin{document}

\maketitle
\thispagestyle{empty} 

\begin{abstract}
In this paper, we first present a method to autonomously detect helipads in real time. Our method does not rely on any machine-learning methods and as such is applicable in real-time on the computational capabilities of an average quad-rotor. After initial detection, we use \textbf{image tracking methods} to reduce the computational resource requirement further. Once the tracking starts our modified IBVS\cite{chaumette2006visual}(Image-Based Visual Servoing) method starts publishing velocity to guide the quad-rotor onto the helipad. The \textbf{modified IBVS} scheme is designed for the four degrees-of-freedom of a quad-rotor and can land the quad-rotor in a specific orientation.
\end{abstract}

\section{Introduction} \label{section:introduction}
Across the world, there is growing interest in the use of quad-rotors for delivery, survey, and emergency response. As the number of these quad-rotors multiply, it will become very time consuming and expensive for human operators to guide these to the intended location, drop off at exact points and then return. As a part of an autonomous ecosystem, it becomes essential that these quad-rotors can land in very precise location with precise orientation for purposes such as re-charging and loading of shipments. \textbf{Precise autonomous landing} is one of the hardest tasks during autonomous navigation of quad-rotors. In certain conditions, the quad-rotor might even be required to land on moving targets. 

The alternate implementations of this task, rely on simple PID controllers\cite{bi2013implementation, saripalli2002vision} which need to be tuned to every drone, are very error prone and orientation invariant. Furthermore, some of these even reconstruct the 3D pose\cite{3dpose} of the helipad which makes it computationally very expensive. 

Our method is based on a feedback loop where we take in the input from a monocular camera mounted at the bottom of the drone and a depth value using SONAR or 1-D LIDAR to calculate the error between the current view of the helipad and the final intended view. After the error is less than a preset threshold, the drone is ordered to land. 

The paper is divided into 7 sections. In section 2 we introduce our method for detecting the helipad initially. Section 3 demonstrates the extraction of image features from the helipad's image to be used for visual servoing. In section 4, we describe how we handle tracking of the helipad while the quad-rotor moves. Section 5 consists of the visual servoing scheme used to find the target velocities. Finally, in section 6 we show the performance of our method on a Parrot Bebop 2  while section 7 concludes the paper.

\begin{figure}[hbt]
\centering
\includegraphics[width=0.8\columnwidth]{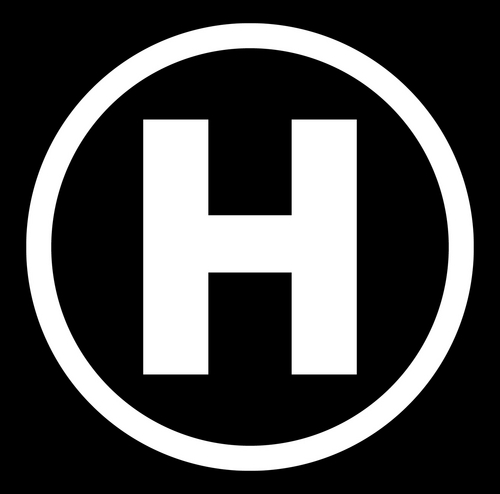}
\caption{The helipad design we used for our testing}
\label{figure:helipad}
\end{figure}

\section{Detection}
\subsection{Overall scheme}
The detection process consists of finding candidate helipads and then eliminating these based on properties of helipads. If a candidate helipad passes all of these elimination tests we move on to the next stage of tracking the helipad.

The properties of a helipad that we rely on during this paper are -
\begin{enumerate}
    \item A bold circle surrounding the H
    \item Presence of H in a bright color inside this circle against a dark background
    \item H is centered at the center of the circle
    \item Intersection of diagonals of H at the center of the circle
\end{enumerate}

Refer figure \ref{figure:helipad} for an example of modern helipad designs.

\subsection{Candidate Detection}
The image is converted to grayscale if it's color. Then we use adaptive thresholding\cite{adaptivethres} to get a binary image. In the binary image, we use Hough transformation\cite{duda1972use, hough1962method} to find the circular regions passing the first criteria. Only circles which get above a certain decided number of votes are considered. These candidates pass to the next stage

\subsection{H Extraction}
Based on the second property, the candidate region is converted to a binary image by thresholding. Based on our experimentation, we found it useful to resize the candidate region image to 228 px x 228 px for further processing as corner detection works optimally in a certain size range. After this, the largest connected component in this image is preserved while the rest of the image is converted into a black region. Finally, the H is smoothened by approximating and then redrawing the contours with blur to prevent false corner detection.

\subsection{Elimination Checks}
At this stage, we have many false positives identifying as potential helipad regions. We use three tests to eliminate these false positives. None of these tests are precise and as a consequence, we have added error ranges based on experiments. 

On the extracted H we find 12 corners using Shi-Tomasi\cite{shi1993good} Corner detection. The process of corner detection is explained in more details in section \ref{section:corner}. From these 12 corners we check whether the midpoints of the 4 outermost points coincide with the center. This is done by calculating the distance from the intersection point to the center of the circle. The distance is then divided by the radius of the circle. This ratio is then checked to be less than 8.25\%.

In the next check, the distance from centroid of H and center of the circle is taken. Similar to the previous one, it is divided by radius of the circle and then checked to be less than 8.25\%. Finally, we check the ratio of the area of H to the area enclosed by the circle. This ratio should be in-between 0.2 to 0.4. If a candidate fails at any of these checks, it is removed from further processing. 

The region which passes these checks is considered to enclose the helipad and is sent to the next stage of this algorithm.

\section{Corner Extraction} \label{section:corner}

\begin{figure}[hbt]
\centering
\includegraphics[width=0.6\columnwidth]{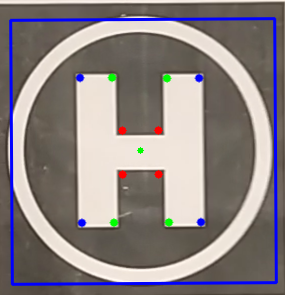}
\caption{Blue box represents tracking region while the corners are classified as explain in section 3}
\label{figure:classified}
\end{figure}

After a binary image is extracted consisting of only the smooth H, we use Shi-Tomasi\cite{shi1993good} to find out the top 12 quality points. These points are then categorized into 3 groups of 4 points each belonging to the outermost H border, the inner H border and the rectangle at the center. See figure \ref{figure:classified} for classified points. 

The classification is done using the idea that the outermost pixels colored blue will have the largest area. When this group is removed the group of pixels colored green will have the largest area, and finally the group of pixels colored as red. This is done by taking first $\binom{12}{4}$ combinations of points and finding the combination with the largest convex area. Then $\binom{8}{4}$ combinations are tested for the next largest area which is assigned to the inner quadrilateral. The last remaining 4 points are allocated to the third group.

\begin{figure}[hbt]
\centering
\includegraphics[width=0.6\columnwidth]{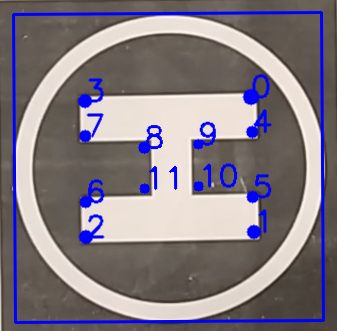}
\caption{Labelled corners inside a helipad}
\label{figure:labelled}
\end{figure}

Inside the groups, the angle of the point relative to the centroid of the helipad is found in the domain of $-{\pi}$ to $+{\pi}$ and then sorted in descending order. The points inside a group are then ordered on the basis of their location in the sorted order. However, if the distance between the first two points is greater than the distance between the first and last point, this order is cyclically shifted by 1. Note that the previous rule is inverted in case of the last group as the longer side of the last group lies along the shorter side of the other two groups of points. Finally, all of these groups are concatenated and the points are ordered 0 to 11 for use during visual servoing. An example of labeled points is figure \ref{figure:labelled}.

\section{Tracking}

The detection process because of its complexity is only run once. After the helipad has been detected we use Median Flow tracker\cite{kalal2010forward} to track the region. Median Flow tracker proved to be fast enough and adapted well to changes in size and orientation of the helipad as the drone moved. 

The tracker, however, at times failed to register the loss of proper tracking. To make sure that the tracker is on the correct region we run only the area check from section 1 at each frame. If at anytime the tracker reports a tracking failure or the area check fails, the tracking segment exits and control goes back to the detection module.

While the tracking segment is active, key points are detected using section \ref{section:corner}. These 12 points are then fed into the IBVS system to retrieve velocities to be published to the quad-rotor.

\section{IBVS}

We use the same definitions for all symbols from paper \cite{chaumette2006visual}. Starting from equation (8), we modify the initial equation from (8) to (8B) by assuming $\omega_x$ and  $\omega_y$ to equal zero at all times. This assumption, while only true during hover, is justified as $\omega_x$ and  $\omega_y$ are proportional to the third derivative of position and as such have a very small impact on the control of the quad-rotor. Furthermore, the output linear velocities can be capped at a maximum value to reinforce this assumption.

\begin{equation}
    \left\{
        \begin{array}{l}
            \dot{x} = \dot{X}/Z - X\dot{Z}/Z^{2} = (\dot{X} - x\dot{Z})/Z  \\*
            \dot{y} = \dot{Y}/Z - Y\dot{Z}/Z^{2} = (\dot{Y} - y\dot{Z})/Z 
        \end{array}
    \right. \tag{7}
\end{equation}
\begin{equation}
    \dot{X} = -\nu_c - \omega_c \times X \iff
        \left\{
            \begin{array}{l}
                \dot{X} = -\nu_x - \omega_yZ + \omega_zY \\*
                \dot{Y} = -\nu_y - \omega_zX + \omega_xZ \\*
                \dot{Z} = -\nu_z - \omega_xY + \omega_yX
            \end{array}
         \right. \tag{8}
\end{equation}

\begin{equation}
    \dot{X} = -\nu_c - \omega_c \times X \iff
        \left\{
            \begin{array}{l}
                \dot{X} = -\nu_x + \omega_zY \\*
                \dot{Y} = -\nu_y - \omega_zX \\*
                \dot{Z} = -\nu_z
            \end{array}
         \right. \tag{8B}
\end{equation}

Using (8B) with (7) from the previously cited paper we get,

\begin{equation}
    \left\{
        \begin{array}{l}
            \dot{x} = -\nu_x/Z + x\nu_z/Z + y\omega_z\\
            \dot{y} = -\nu_y/Z + y\nu_z/Z - x\omega_z
        \end{array}
    \right. \tag{9B}
\end{equation}

Equation (10) remains unchanged

\begin{equation}
    \dot{x} = L_xv_c \tag{10}
\end{equation}

The new interaction matrix is of the form 
\begin{equation}
    L_x =
        \begin{bmatrix}
            \frac{-1}{Z} & 0 & \frac{x}{Z} & y \\
            0 & \frac{-1}{Z} & \frac{y}{Z} & -x
        \end{bmatrix} \tag{11B}
\end{equation}

As we have a total of 12 feature points, our final interaction matrix is ${L_x} \in \mathrm{}{R^{24\times4}}$

The error metric that w use 

\section{Results}

\begin{figure}[hbt]
\centering
\includegraphics[width=0.9\columnwidth]{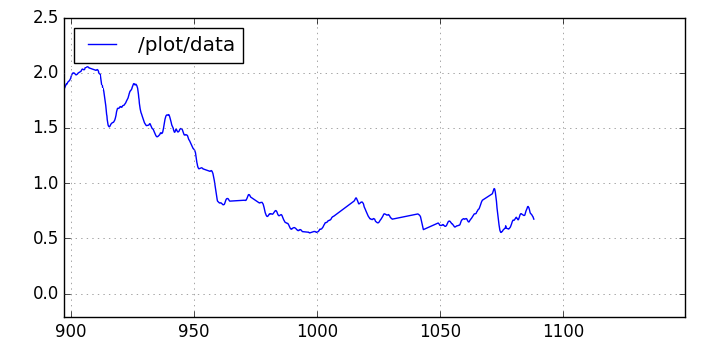}
\caption{Time variation of error for indoors test during visual servoing.}
\label{figure:egraph1}
\end{figure}

\begin{figure}[hbt]
\centering
\includegraphics[width=0.9\columnwidth]{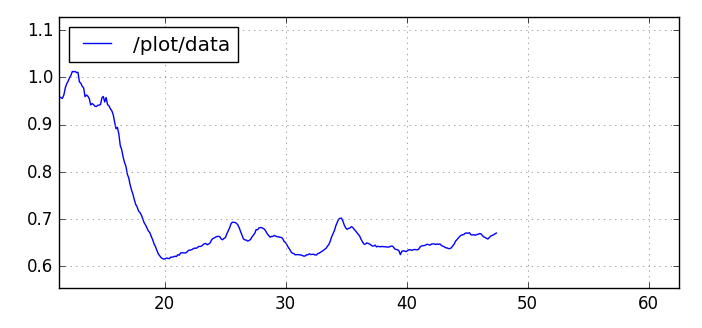}
\caption{Time variation of error for outdoors daytime during visual servoing.}
\label{figure:egraph2}
\end{figure}

\begin{figure}[hbt]
\centering
\includegraphics[width=0.9\columnwidth]{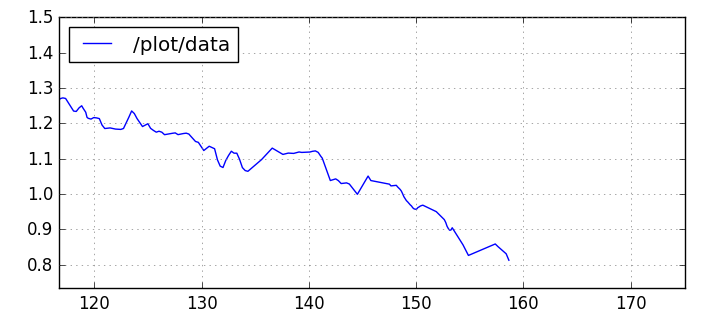}
\caption{Time variation of error for outdoors low-light during visual servoing.}
\label{figure:egraph3}
\end{figure}
For the testing of our algorithm, we used a Parrot Bebop 2\cite{parrot2016parrot} quad-rotor connected to a PC running the bebop autonomy driver\cite{monajjemi2015bebop} on the ROS system to eliminate any hardware limitations posed by a custom made drone. We calibrated the camera using camera\_calibration\cite{bowmanros} from ROS. The reference image features based on the final position of the quad-rotor just above helipad was saved. 

The live feed was taken using the high field of view camera of bebop, and the region of interest was the vertically downward facing. Also using the ROS drivers, the linear and angular velocity was fed to the interface. The coordinate axis of the IBVS and that of bebop were not the same and thus required transformations were made to match them. The IBVS's output velocities were also scaled to match the range for Bebop's controls. The Bebop does not have a stereo camera; however, it has a sonar through which the bebop autonomy driver returns depth value. We approximate the depth value for all the points to equal this value returned by bebop autonomy. While not ideal, it worked good in practice.

\begin{figure}[hbt]
\centering
\includegraphics[width=0.59\columnwidth]{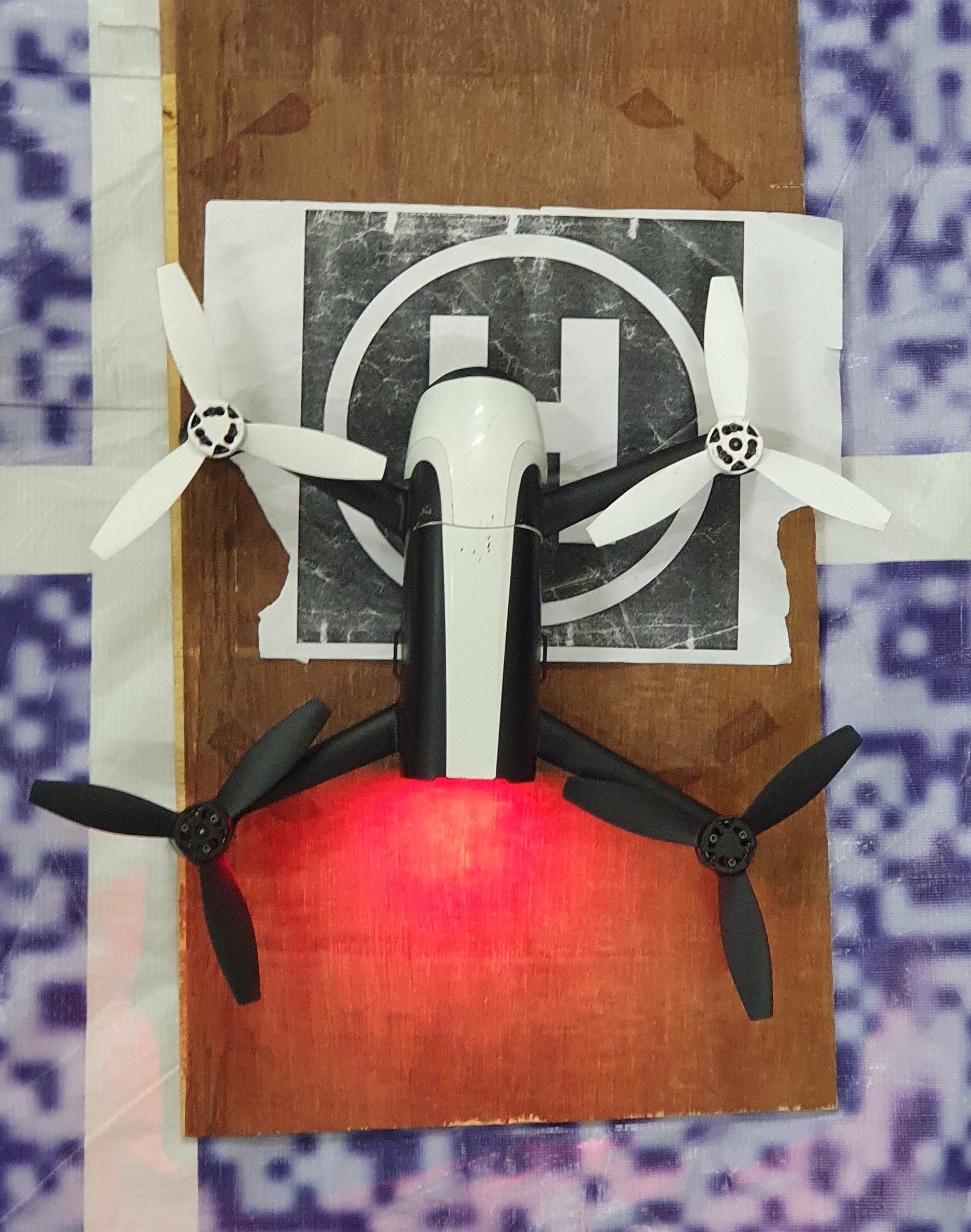}
\caption{Configuration after an indoors land test}
\label{figure:final1}
\end{figure}

\begin{figure}[hbt]
\centering
\includegraphics[width=0.57\columnwidth]{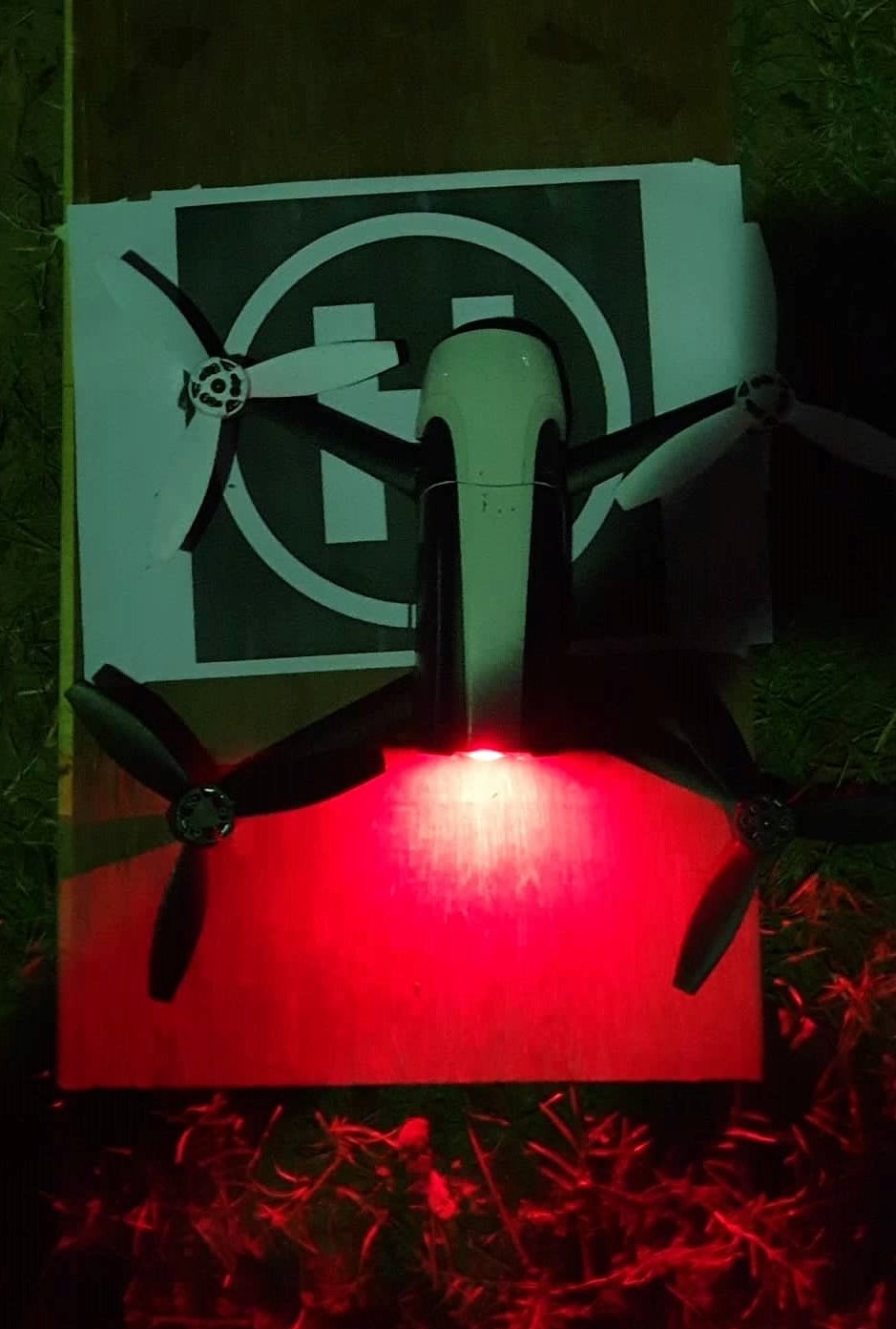}
\caption{Configuration after an outdoors low-light land test}
\label{figure:final2}
\end{figure}

The drone took off and was manually guided to a high altitude position from which it could view the helipad after which our algorithm took control of the drone. Then, we let it guide the drone till error reached a steady state. Once, the error stabilized, the land command was issued and the drone landed. Examples of final landing position are described by figures \ref{figure:final1},\ref{figure:final2}. Note that the center of the quad-rotor is taken as the camera which is in the front part of the drone for Bebop.

The error is calculated by taking the norm of the error vector as described in equation (1) from \cite{chaumette2006visual}. Figures \ref{figure:egraph1},\ref{figure:egraph2},\ref{figure:egraph3} show the variation of error versus time.

The source code for this test is available at \url{https://github.com/archit120/bebop_precision_landing}

\section {CONCLUSION}

We have presented the design and implementation of a real-time vision-based algorithm for autonomously landing a quad-rotor. The algorithm adopts simple, fast, and robust methods proven to work under various circumstances as its building blocks.

The detection method is based on simple geometrical features present in every helipad to land a drone with extreme precision in translation and rotation. Based on our experiments, we find this algorithm works even for moving targets and landing at oblique orientations. Finally, as we use visual servoing instead of PID controllers\cite{bouabdallah2004pid}, no fine tuning is required for different systems.

\section* {Acknowledgement}
This research was supported by \textbf{Aerial Robotics Lab, IIT Kharagpur}. We are members of this lab and we thank the management for providing resources that allowed the completion of our work. We also thank our colleagues from Aerial Robotics Lab, IIT Kharagpur who provided insight and expertise that greatly assisted the research.

We would also like to acknowledge the drone \textbf{Parrot Bebop 2\cite{parrot2016parrot}} which we used for our testing for its stability and sensors which enabled us to perform our tests.

\bibliographystyle{unsrt}
\bibliography{imav_template}

\appendix
\newcommand{\appsection}[1]{\let\oldthesection\thesection
  \renewcommand{\thesection}{Appendix \oldthesection:}
  \section{#1}\let\thesection\oldthesection}

\end{document}